\documentclass[letterpaper, 10pt, conference]{ieeeconf}  

\IEEEoverridecommandlockouts                              

\usepackage{algorithm}
\usepackage[noend]{algpseudocode}

\usepackage{threeparttable}
\usepackage{multirow}
\usepackage{array}
\usepackage{graphicx}
\usepackage{textcomp}
\usepackage{xcolor}
\usepackage{lipsum}
\usepackage{float}

\usepackage{hyperref}
\hypersetup{
    colorlinks=true,
    linkcolor=black,
    filecolor=magenta,  
    citecolor=black,
    urlcolor=cyan,
    pdftitle={Pour me a drink},
    pdfpagemode=FullScreen,
    }

\usepackage[
    backend=biber,
    natbib=true,
    sorting=none,
    style=ieee,
    citestyle=numeric,
] {biblatex}

\newlength{\IEEEbibitemsep}
\title{\LARGE \bf
Pour me a drink: Robotic Precision Pouring Carbonated Beverages into Transparent Containers
}

\author{Feiya Zhu$^{1}$, Shuo Hu$^{*1}$, Letian Leng$^{*1}$, Alison Bartsch$^{1}$, Abraham George$^{1}$, Amir Barati Farimani$^{1}$
\thanks{$^{*}$These authors made an equal contribution to this work}
\thanks{$^{1}$Feiya Zhu, Shuo Hu, Letian Leng, Alison Bartsch, Abraham George and Amir Barati Farimani are with the Department of Mechanical Engineering, Carnegie Mellon University, 5000 Forbes Avenue, Pittsburgh, 15213, PA, USA
        {\tt\small \{feiyaz, shuohu, lleng, abartsch, aigeorge, afariman\}@andrew.cmu.edu}}%
}

\begin{document}

\maketitle
\thispagestyle{empty}
\pagestyle{empty}

\begin{abstract}

With the growing emphasis on the development and integration of service robots within household environments, we will need to endow robots with the ability to reliably pour a variety of liquids. However, liquid handling and pouring is a challenging task due to the complex dynamics and varying properties of different liquids, the exacting precision required to prevent spills and ensure accurate pouring, and the necessity for robots to adapt seamlessly to a multitude of containers in real-world scenarios. In response to these challenges, we propose a novel autonomous robotics pipeline that empowers robots to execute precision pouring tasks, encompassing both carbonated and non-carbonated liquids, as well as opaque and transparent liquids, into a variety of transparent containers. Our proposed approach maximizes the potential of RGB input alone, achieving zero-shot capability by harnessing existing pre-trained vision segmentation models. This eliminates the need for additional data collection, manual image annotations, or extensive training. Furthermore, our work integrates ChatGPT, facilitating seamless interaction between individuals without prior expertise in robotics and our pouring pipeline, this integration enables users to effortlessly request and execute pouring actions. Our experiments demonstrate the pipeline's capability to successfully pour a diverse range of carbonated and non-carbonated beverages into containers of varying sizes, relying solely on visual input. For comprehensive demonstrations, please refer to the videos available on our project page: {\tt\small \url{  https://sites.google.com/andrew.cmu.edu/robotcokepouring/}}

\end{abstract}

    \begin{figure}[thpb]
        \centering
        \includegraphics[width=0.4\textwidth]{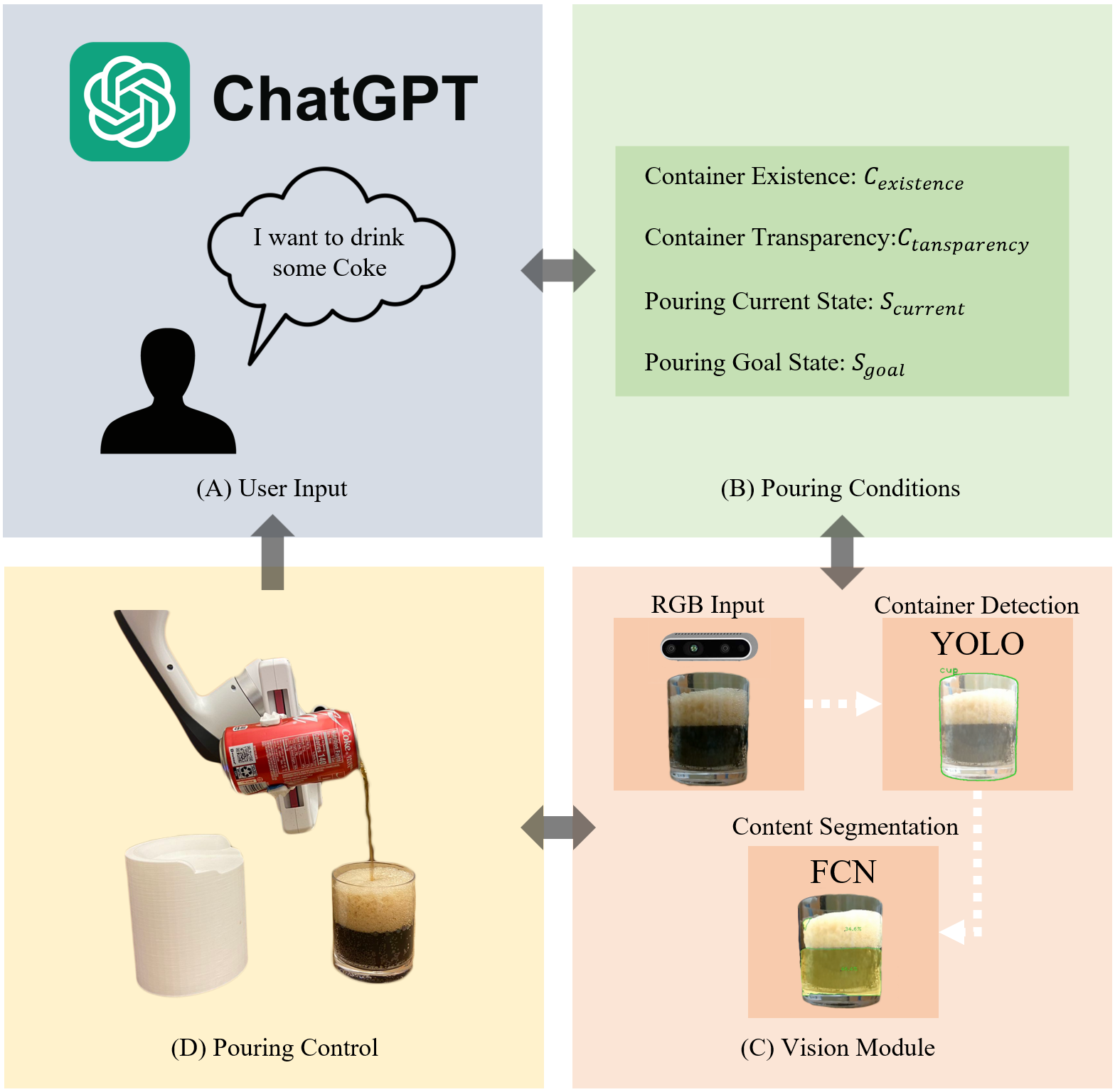}
        \caption{Overview of Our Robotic Pouring Pipeline: (A) Generating pouring goals via user interaction with ChatGPT. (B) Cross-checking predefined pouring conditions with the vision module. (C) Vision module for container and content segmentation. (D) Robotic pouring execution.}
        \label{fig:pipelineOverview}
    \end{figure}

\section{INTRODUCTION}

Autonomous liquid pouring by robots holds great potential for streamlining various household tasks, notably beverage preparation. Nevertheless, achieving the precise and rapid pouring of carbonated beverages presents a formidable challenge, especially in the context of robotics. This difficulty arises from the dynamic nature and diverse characteristics of carbonated drinks, where the rapid and unpredictable expansion of bubbles and foam during pouring becomes a critical factor that directly influences the attainment of the desired pouring outcome. Additionally, the varying levels of transparency among different types of carbonated beverages, such as Coke and Sprite, introduce an additional layer of complexity, demanding precise control throughout the pouring process.

Most previous research on robotic pouring has predominantly centered around non-carbonated liquids. These studies have employed various approaches for detecting liquids inside containers. Some rely on visual input from cameras \cite{schenck_visual_2017, kennedy_autonomous_2019, burns_look_2022, narasimhan_self-supervised_2022, dong_precision_2019, do_accurate_2018}. Others \cite{burns_look_2022, liang_robust_2020} use microphones to capture pouring sounds. Additionally, certain methods incorporate tactile or force sensors \cite{liang_robust_2020, matl_haptic_2019}. A subset of techniques opts for fluid simulation to perform pouring tasks \cite{li_graph_2022, babaians_pournet_2022, xian_fluidlab_2023}. Nonetheless, these approaches exhibit various limitations. Some necessitate the collection of extensive training datasets, while others depend on specialized environmental equipment, such as scales placed beneath the target container. Furthermore, many of these methods are designed exclusively for single-phase fluids such as water. 

To address these limitations and challenges, we leveraged existing pre-trained vision detection and segmentation models, as they successfully segment both transparent containers and fluids sufficiently for pouring tasks and do not require any further data collection. In this work, we present a vision-based pipeline that empowers robots to proficiently pour precise quantities of both carbonated and non-carbonated liquids into transparent containers commonly encountered in domestic settings. This is achieved through a fully autonomous approach that capitalizes on the capabilities of vision models YOLO \cite{redmon_you_2016,jocher_yolo_2023} and an FCN-based network \cite{long_fully_2015,eppel_computer_2021,ronneberger_u-net_2015,chen_rethinking_2017}. Additionally, our pipeline incorporates a simple PID controller augmented by supplementary control policies, as well as a physics-based model for volume estimation.

Our approach pioneers the direct integration of well-established pre-trained segmentation models, originally designed for other applications, into robotic pouring tasks. Notably, our pipeline excels in real-time pouring control and volume estimation, representing a significant advancement in liquid pouring automation. The key contributions of this paper include:

\begin{itemize}
\item \textbf{Autonomous:} Our modular framework handles real-time pouring tasks with both non-carbonated and carbonated liquids, regardless of transparency, and without supervision. 

\item \textbf{Zero-shot:} We rely solely on pre-trained models, eliminating data collection and training requirements, and simplifying the process for researchers and users.

\item \textbf{User-friendly:} Integration with ChatGPT allows users with no prior knowledge to efficiently control pouring actions, showcasing the potential of large language models in streamlining robotic system utilization.
\end{itemize}

\section{RELATED WORKS}

\textbf{Liquid Perception:}
Achieving precise robotic pouring into transparent containers necessitates accurate detection of these containers. Past research has addressed this challenge using diverse strategies, including object detection and segmentation models \cite{jocher_yolo_2023, eppel_computer_2021, xie_segmenting_2020, xie_segmenting_2021}. These methods leverage deep neural networks, often trained on extensive annotated datasets. Alternatively, some techniques employ RGB-D cameras to estimate shape and depth \cite{zhu_rgb-d_2021, sajjan_cleargrasp_2019}. Others predict object poses using 3D key-points derived from stereo input \cite{liu_keypose_2020}. Once the target container is localized, determining the liquid's status inside becomes essential. Vision-based approaches are prevalent in achieving liquid perception for robotic pouring. For example, \cite{schenck_visual_2017} used a CNN-LSTM network trained on RGB-D images, supplemented with thermal camera data to generate ground truth labels. \cite{do_accurate_2018} incorporated a Kalman filter, utilizing depth information from an RGB-D camera to detect and track both opaque and transparent liquid levels. More recently, \cite{narasimhan_self-supervised_2022} introduced a technique involving a dataset of colored liquid, applying background subtraction to generate liquid ground truth from colored liquid pixels, and the dataset was used to train a UNet network for liquid segmentation.

\textbf{Liquid Manipulation:}
The manipulation of liquids by robotic systems, especially in household scenarios like kitchens \cite{dikshit_robochop_2023}, has attracted significant attention. Precise and spill-free pouring is a crucial aspect of liquid manipulation. Previous research has primarily focused on precision pouring \cite{schenck_visual_2017, kennedy_autonomous_2019, dong_precision_2019,do_accurate_2018,liang_robust_2020,  babaians_pournet_2022}, aiming to transfer specific volumes of fluids into designated containers. Some studies \cite{do_accurate_2018, babaians_pournet_2022} have conducted pouring experiments with various fluids, including water, juice, Coke, and beer. However, \cite{babaians_pournet_2022} acknowledges the challenges posed by effervescent fluids like beer and their inability to conduct pouring experiments with beer due to complications arising from foam generation in accurate simulations and real-world pouring scenarios. Additionally, certain studies, such as \cite{do_accurate_2018}, have focused on pouring Coke but employed point cloud-based liquid detection, which does not account for foam formation, treating it as a transparent liquid without recognizing its unique characteristics.

\textbf{Volume estimation:}
Some studies have primarily focused on liquid height detection rather than actual volumetric measurements \cite{narasimhan_self-supervised_2022, do_accurate_2018}. However, precise liquid volume estimation is crucial for a wide range of robotic pouring tasks, especially those involving the mixing of liquid ingredients for culinary purposes. Various methods have been developed to accurately estimate liquid volume in containers. For instance, in \cite{schenck_visual_2017, schenck_towards_2016}, RGB image inputs, combined with thermal images, are used to generate a probability distribution over liquid volume. \cite{burns_look_2022} introduces a comprehensive multi-sensory pouring dataset that includes RGB images and pouring audio. \cite{dong_precision_2019} focuses on the geometric attributes of both target and pouring containers. Similarly, \cite{kennedy_autonomous_2019} concentrates on the geometric properties of pouring containers and incorporates weight detection to gauge the fluid's weight. Additionally, \cite{liang_robust_2020, matl_haptic_2019, piacenza_pouring_2022} utilize force and torque information to improve volumetric estimation accuracy. Notably, \cite{matl_haptic_2019, reyes-montiel_geometric_2022} consider pouring container geometry as an influential factor.

\textbf{Large Language Models in Robotics:}
Natural Language Processing (NLP) has become vital for human-robot communication, with the transformer architecture introduced \cite{vaswani_attention_2023}, which serves as the foundation for numerous Large Language Models (LLMs). Microsoft recently unveiled a simulation tool that integrates ChatGPT into Microsoft Airsim \cite{shah_airsim_2017, vemprala_chatgpt_2023}, exemplifying a seamless natural language interface between users and robots. In our work, we leverage ChatGPT to customize user inputs for robot control, facilitating pouring based on natural language instructions. 

\begin{algorithm}
\caption{Robotic Pouring Pipeline}
\begin{algorithmic}[1]
\State Initialize vision model: $\hat{S}_{I}$, ($C_{E}$, $C_{T}$, $l_{I}$) 
\Statex {$C_{E}$ for container existence, $C_{T}$ for container transparency, $l_{I}$ for initial liquid status.}
\State Initialize ChatGPT loop: ($C_{E}$, $C_{T}$, $l_{I}$)
\State ChatGPT $\gets$ True

\While{ChatGPT}
    \If{$\neg C_{E}$}
        \State Prompt user: Place a cup, $C_{E}$ $\gets$ \(\hat{S}_{t}\)
    \ElsIf{$\neg C_{T}$}
        \State Prompt user: Please use a transparent cup
        \State $C_{T}$ $\gets$ \(\hat{S}_{t}\)
    \Else
        \State Get user input: Target liquid level, $l_{target}$
        \State Pouring $\gets$ True
        \While{Pouring}
            \State Perform robotic pouring: $\hat{S}_{t}$
            \State $f_t \gets$ Foam status from $\hat{S}_{t}$
            \State $l_t \gets$ Liquid status from $\hat{S}_{t}$
            \If{($l_t$ + $f_t$) $>$ $l_{target}$}
                \State Observe and wait for foam to settle
            \Else 
                \State Continue pouring
                \State Pouring $\gets$ False if $l_t$ $\leq$ $l_c$ 
            \EndIf
        \EndWhile
        \State Ensure termination of pouring action
        \State Display pouring completion message
        \State ChatGPT $\gets$ False \Comment{Exit ChatGPT loop}
    \EndIf
\EndWhile
\end{algorithmic}
\end{algorithm}

\begin{figure}
\begin{center}
  \includegraphics[width=0.5\textwidth, height=9cm]{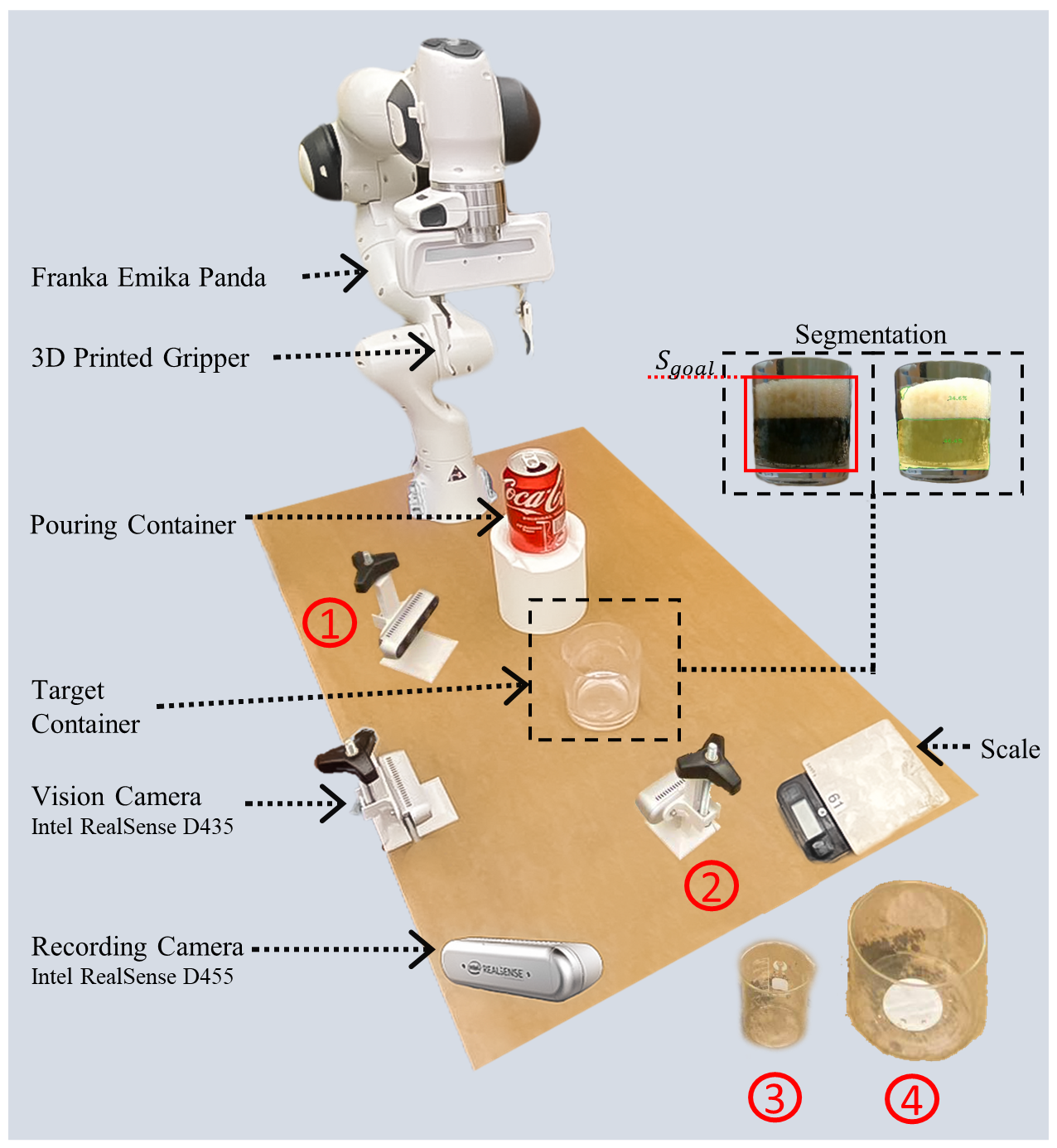}
  \caption{A visualization of our experiment setup. An Intel RealSense Camera captures RGB images for vision segmentation, with another camera recording the pouring process. A 3D-printed support assists the gripper in picking up the pouring container (Coke can). A scale measures liquid weight for volume estimation reference. Numbers 1 and 2 indicate additional camera angles, while 3 and 4 represent diverse target containers used in experiments.}
  \label{fig:ExperimentalSetUp}
  \end{center}
\end{figure}

\section{METHODOLOGY}
In this study, we aim to achieve precise pouring for a wide range of liquids, including carbonated and non-carbonated beverages. We introduce an autonomous robotic pouring pipeline that operates solely on a single RGB input, as illustrated in Fig.\ref{fig:pipelineOverview}. 

\textbf{Vision Module:} 
To analyze the internal state of transparent containers, our visual module employs image segmentation to transform a single RGB image input into analytically interpretable representations. Many standard segmentation tasks typically require the creation of large image datasets with detailed human annotations. However, our approach builds upon pre-trained segmentation models obtained from \cite{jocher_yolo_2023, eppel_computer_2021}, eliminating the need for data collection and manual annotations.
We initially used a YOLO v8 segmentation model from Ultralytics \cite{jocher_yolo_2023} to detect transparent containers in our pouring scene. This YOLO model was pre-trained on the COCO dataset \cite{lin_microsoft_2015}, which includes 118,287 images with 80 object categories.It proved more than adequate for detecting target containers in our method. For an RGB input, this model provides a bounding box, a segmentation mask, pose keypoints, and class probabilities. For the purpose of this work, we found that the bounding box, segmentation masks, and class probabilities are sufficient for detecting and segmenting the target container. The pose key-points are not considered in this task but could be applied in future robotic pouring tasks, such as 3D pose estimation \cite{liu_keypose_2020}. 
While the YOLO model can detect the existence of target containers ($C_{existence}$), its mask output does not include container properties ($C_{transparency}$) or the content status inside the container. This limitation hinders the required information for visually controlling the robotic pouring and performing content segmentation within the container. Therefore, we incorporated another pre-trained segmentation model from \cite{eppel_computer_2021}. This model was pre-trained on the LabPics dataset, which contains 5,007 training images, including annotations on container types, properties, and corresponding materials (liquid, suspension, foam). We attempted to augment this FCN-based model by combining it with their prior Vector-LabPics dataset \cite{eppel_computer_2020}. However, providing more image data did not show a significant improvement for our pouring segmentation. Still, this addition could prove beneficial for more complex environmental settings. Therefore, we do no further fine-tuning with the pre-trained model for fluid and foam segmentation. 
With these two pre-trained segmentation models at our disposal, the YOLO model first outputs the detection results for target containers ($C_{existence}$). If there is a target container in the pouring scene, the corresponding segmentation mask of the container from the YOLO model is passed to the content segmentation model. The container mask from the YOLO model serves as a region of interest (ROI) input for content segmentation. Consequently, the content segmentation provides the container properties (transparency) ($C_{transparency}$) and the content states ($\hat{S}_{current}$), where content states include liquid and foam information for this work.

\textbf{Robotic Pouring Control:} 
In this work, we employ a vision module for real-time monitoring of the target container's content status during pouring. Our pouring control strategy involves regulating the end-effector's y-axis rotation angle via a PID controller, complemented by additional control policies, facilitating precise and smooth pouring. To enhance clarity and manage complexity, we segment the pouring control into three distinct stages: pre-pouring, intra-pouring, and post-pouring.
In the pre-pouring stage, the robot retrieves the source container, and we assume its availability. The vision module assesses key target container parameters, including its existence ($C_{E}$), transparency ($C_{T}$), and current pouring status ($\hat{S}_{I}$). Once these conditions are satisfied, the robot transitions to the intra-pouring stage. During the intra-pouring stage, the end-effector initiates a predefined rotation of the source container, specifically 55 degrees from vertical. This approach minimizes visual segmentation errors and addresses challenges related to initial liquid detection, effectively preventing spills shown in Fig.\ref{fig:RobotPouring} (C)(c). We term this process the Initial Pour. The PID controller utilizes the discrepancy between the target and current liquid percentages as input errors, dynamically adjusting the end-effector's rotation angle. Pouring continues until one of three conditions is met: the target liquid percentage is achieved, the target container reaches its capacity, or the pouring container is depleted of liquid.
In the post-pouring stage, particularly critical for carbonated beverages, the robot employs a wait-to-observe policy. Upon reaching the target liquid percentage for the first time, the robot halts end-effector rotation and enters a holding position. The pouring container remains tilted and ready for further pouring, with the duration of this phase contingent on the settling rate of foam. The robot subsequently determines whether to resume pouring or conclude the pouring task.

\textbf{Volume Estimation:}
In volume estimation, our approach draws inspiration from prior works such as \cite{kennedy_autonomous_2019, dong_precision_2019, matl_haptic_2019, reyes-montiel_geometric_2022}. Given the flexibility of our method to adapt to various transparent containers, we have chosen to focus on a practical, everyday example: the Coke can. Additionally, our approach integrates insights from \cite{liang_robust_2020, matl_haptic_2019, piacenza_pouring_2022}, leveraging the haptic information, specifically joint rotation data, provided by the robot arm. This eliminates the need for additional sensors, simplifying the volume estimation process. In our volume estimation approach, we make several key assumptions to streamline the process: we rely on the positional data from the end-effector wrist joint's rotation, obtained from the Franka robot arm, as our primary source of information. Our experiments exclusively use Coke cans as the pouring containers. We treat these cans as right circular cylinders shown in Fig.\ref{fig:RobotPouring} (C)(d), assuming that their wall thickness is negligible. This simplifies volume calculations based on their external geometry. The robot consistently selects the pouring angle at the same height along the length of the pouring container. This uniformity further simplifies our volume estimation process.

\subsubsection{Analytic Geometric Volume Estimation}
Referring to Fig.\ref{fig:ExperimentalSetUp} (D), we establish certain volume parameters within the pouring process: $V_{total}$ represents the total volume of liquid contained within the pouring container. $V_{offset}$ signifies the quantity of liquid that travels from the pouring container before reaching the target container. This occurs when the visual module detects that the liquid level has reached the target level and commands the robot to halt the pouring action. Some liquid remains suspended in the air during this process. $V_{pour}$ denotes the volume of liquid situated above the container's opening edge at a given rotating angle $\theta_{pour}$. This measurement includes the volume $V_{offset}$. $V_{stay}$ represents the volume of liquid below the container's opening edge at the same rotating angle $\theta_{pour}$.

\begin{equation}
V_{total} = V_{pour} + V_{stay}
\end{equation}

\begin{equation}
V_{pour} = V_{liquid} + V_{offset}
\end{equation}

Our approach to volume estimation draws inspiration from the work in \cite{matl_haptic_2019}. This method, which is tailored for cylindrical containers, estimates the pouring angle by pouring a specific volume from the pouring container. By adapting their pouring approach to our definitions, we establish a relationship between the pouring angle $\theta$ and the poured-out liquid volume $V_{pour}$.

\begin{equation}
\theta = \arg\min_{\theta \in [55^\circ, 90^\circ]} \| V_{pour} - (V_{total} - V_{stay}L\cos{\theta}) \|_2^2 
\end{equation}

Equation (3) illustrates the relationship between $V_{pour}$ and $\theta_{pour}$, enabling us to calculate $\theta_{pour}$ when given $V_{pour}$. However, it's essential to note that volume estimation doesn't directly influence our pouring control process. Instead, we conduct volume estimation after completing the pouring action, utilizing the Franka robot's rotation angle, from which we obtain the necessary data.

\subsubsection{Liquid Weight Volume Estimation}
In geometric volume estimation, assuming the Coke Can's cylindrical shape, we enhance volume estimation with an empirical relationship between poured liquid volume ($V_{pour}$) and rotation angle ($\theta_{pour}$). Experiments measure $V_{pour}$ by weighing the target container post-pouring, assuming constant liquid density ($V = \frac{M}{\rho}$).

We vary the end-effector's angle from $50^\circ$ to $100^\circ$ in $5^\circ$ steps. After each pour, a 20-second pause lets foam settle. Three weight measurements yield an average, from which liquid volume is determined using its density. With a consistent pouring container, volume behavior aligns with rotation angle. We use a degree-3 polynomial regression to establish the $f(x) = c_{0} +c_{1}x^{1}+c_{2}x^{2}+c_{3}x^{3}$ relationship using experimental data.

\begin{figure*}
\begin{center}
  \includegraphics[width=\textwidth]{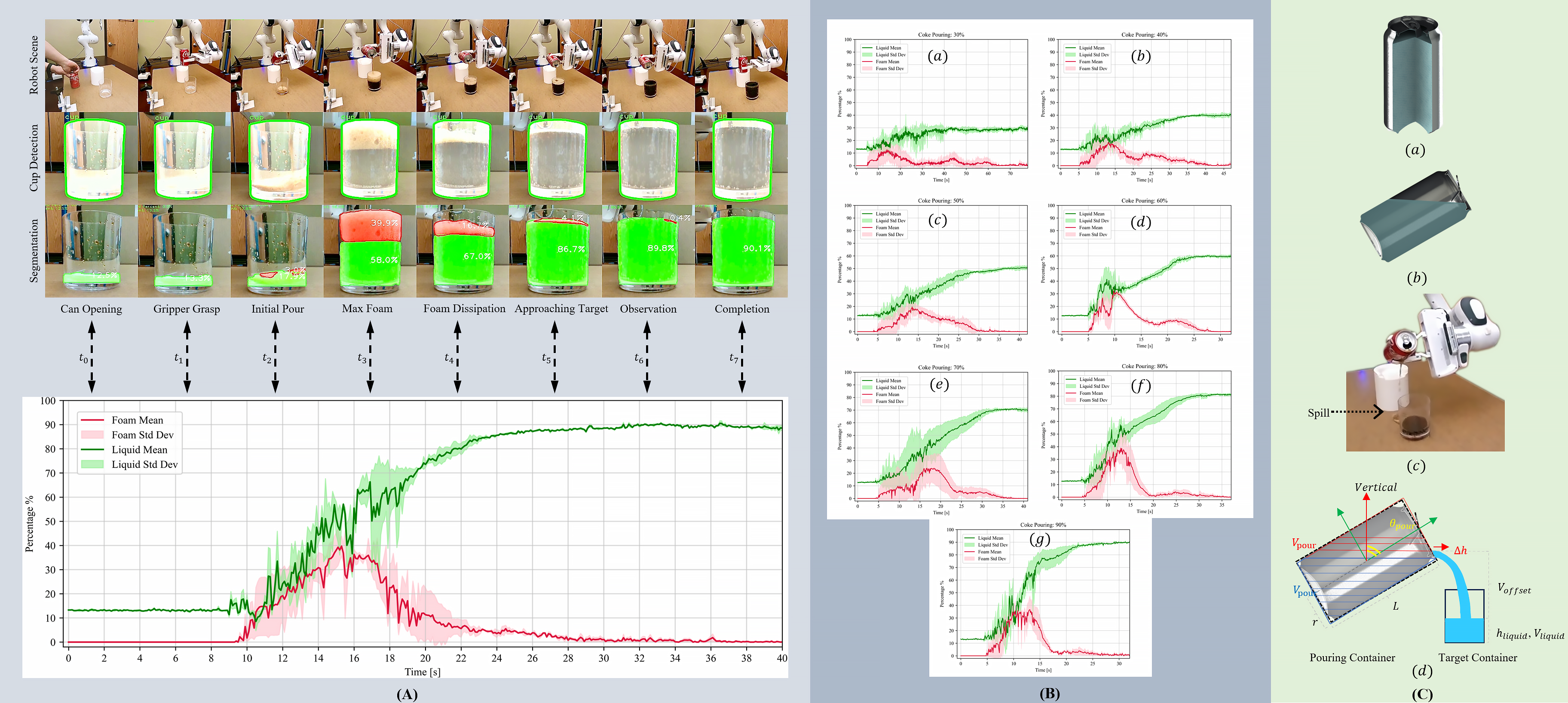}
  \caption{Illustration of robotic pouring at various time-steps and the performance of our Coke pouring pipeline. \textbf{(A)} Visualizes the robotic pouring process at different stages. \textbf{(B)} Demonstrates the performance of our pipeline during Coke pouring, ranging from 30\% (a) to 90\% (b) of liquid in the target container. Red and green lines represent foam and liquid level changes, with shaded areas indicating standard deviations. \textbf{(C)} Features sections showing a full Coke Can (a), the rotating can before liquid egress (b), a pouring spill due to surface tension (c), and a cross-sectional diagram defining variables used for volume estimation (d).}
  \label{fig:RobotPouring}
  \end{center}
\end{figure*}

\textbf{ChatGPT Integration:}
We draw inspiration from the work by \cite{vemprala_chatgpt_2023}, which highlights the effectiveness of ChatGPT in providing a highly intuitive natural language interface between users and robots. This integration presents a zero-shot solution for solving robotics problems through the use of ChatGPT. In our methodology, ChatGPT assumes the role of an interpreter between the user and our pouring pipeline. It possesses the capability to handle ambiguous and ill-defined instructions while also being able to seek clarification from the user when necessary. Algorithm 1 provides a summary of our approach with ChatGPT integration for autonomous pouring.

\section{EXPERIMENTS}

In our experiments, we evaluate the efficacy of our pouring pipeline in a zero-shot scenario. We begin by evaluating its performance in pouring Coke, followed by an examination of its robustness and versatility across various liquids, distinct vision camera positions, and diverse target containers.

\textbf{Experiment Setup:}
Our robotic pouring experiment employed a Franka-Emika Panda 7-DOF robotic arm with a custom 3D-printed gripper for pick-and-pour tasks. The pouring control module was integrated using the Franka robot control framework \cite{zhang_modular_2020}. The robot retrieved the pouring container from a fixed location and positioned it above the target container. An Intel RealSense D435 RGBD camera captured the target container's RGB image at 60 FPS, with a resolution of 640x480 pixels for vision segmentation. Our setup is visually detailed in Fig.\ref{fig:ExperimentalSetUp}. Carbonated beverage pouring included manual can opening before robotic pouring, with each pour using a new can for consistent carbonation.

\textbf{Experiment Design:}
To assess the overall pouring performance of our pipeline, we poured Coke into the same target container at fluid levels ranging from 30\% to 90\% (in 10\% increments). We initiated at 30\% due to initial vision segmentation errors with an empty transparent cup, as shown in Fig.\ref{fig:RobotPouring} (A) before Initial Pour stage. Five trials were conducted for each level, and the average and standard deviation were calculated. Pouring time was measured from the Initial Pour stage to Completion, with weight measurements providing ground truth for volume estimation.

To evaluate the pipeline's robustness and adaptability, shown in Fig.\ref{fig:ExperimentalSetUp}, we extended our experiments to include Mountain Dew (MTN DEW), Sprite, and water as representative beverages. These choices covered a spectrum of properties, such as carbonation, transparency, and color. We also introduced variations by testing in different camera locations with distinct visual backgrounds and using various target containers, each with its unique characteristics.

\section{RESULTS}
In this section, we present the results of our experiments evaluating the performance of our autonomous robotic pouring pipeline. The table provides data on the final fluid level achieved, standard deviation ($\sigma$) of the final fluid level, predicted volume (P.V.), final volume (F.V.), standard deviation ($\sigma$) of predicted volume, and pouring time. Our pipeline consistently demonstrated pouring accuracy, with an average final percentage error below 1\%. For pouring Coke, $\sigma$ remained below 1.5\%, while for MTN DEW, Sprite, and water, $\sigma$ was under 0.5\%, 0.5\%, and 0.2\% respectively. This precision extended to liquid volume estimation, with an average final volume error below 7$ml$ and $\sigma$ below 10$ml$ for Coke. For MTN DEW, the average volume error was below 6$ml$ with $\sigma$ under 5$ml$, while for water, the average error was below 3$ml$ with $\sigma$ under 5$ml$. For Sprite, the volume error was below 8$ml$, and $\sigma$ remained under 4$ml$.

\subsection{Pouring Performance with Coke}
Table \ref{Coke} summarizes the pouring performance of Coke into the target container at various target fluid levels. The results indicate that our pipeline consistently achieved final fluid levels within 1\% of the target, with volume deviations ranging from 2 ml to 7 ml, primarily centered around 4\%. This level of error, considering the total volume of a Coke can (355 ml), is deemed acceptable within the context of pouring time optimization. The optimal performance was observed for target percentages ranging from 60\% to 80\%, with the longest pouring time occurring at a target percentage of 30\%, primarily due to initial segmentation errors close to the target value, shown in Figure \ref{fig:RobotPouring}(A) before Initial Pour stage.

\newcolumntype{C}[1]{>{\centering\arraybackslash}m{#1}}

\begin{table}[h!]
  \centering
  \caption{Pouring performance of the Coke}
  \label{Coke}
  \begin{threeparttable}
    \begin{tabular}{|C{0.82cm} | C{0.82cm} C{0.82cm}| C{0.82cm} C{0.82cm} C{0.82cm} | C{0.82cm}|}
      \hline
      Target & Final & $\sigma$ & P.V.\tnote{1} & F.V.\tnote{2} & $\sigma$ & Time\\
      {\%} & {\%} & {\%} & {$ml$} & {$ml$} & {$ml$} & {$s$} \\
      \hline
      30 &  30.82 & 1.38 & 90.4 & 84.0 & 7.27 & 54.17\\
      40 &  40.30 & 0.74 & 128.0 & 124.0 & 8.64 & 37.04 \\
      50 & 50.68 & 0.58 & 168.9 & 172.8 & 9.43 & 35.02 \\
      60 & 59.48 & 1.09 & 203.7 & 196.8 & 4.35 & 26.44 \\
      70 & 70.82 & 0.78 & 244.9 & 240.0 & 6.29 & 26.19 \\
      80 & 80.78 & 0.64 & 285.4 & 283.8 & 4.66 & 26.96  \\
      90 & 89.62 & 0.53 & 316.4 & 318.4 & 1.36 & 36.0  \\
      \hline
    \end{tabular}
    \begin{tablenotes}
        \item [1] P.V. = Predicted Volume 
        \item [2] F.V. = Final Volume
    \end{tablenotes}
  \end{threeparttable}
\end{table}

\subsection{Pouring Generalization}
To assess the robustness and generalization of our pipeline, we extended our experiments to include different types of beverages, diverse target containers, and varied camera locations.

\subsubsection{Pouring Performance with Different Beverages}
Table \ref{diff beverage} presents the pouring performance for three different beverages: MTN DEW, Sprite, and water. Our system consistently achieved remarkably accurate pouring results across various beverages, with final fluid levels closely matching target percentages (average error < 1\%).  Low volume deviations (standard deviations $\sigma$) further highlight precision; at a 60\% target, the deviation was only 0.22 ml. However, Sprite presented the greatest challenge, with its visual segmentation being easily affected by the bubbles within the beverage. This issue was less prominent in MTN DEW and Coke due to their distinct colors, providing better visual contrast compared to Sprite. Additionally, Water had the shortest average pouring time, primarily because it contains no bubbles to interfere with our vision module. As a result, our pipeline excelled in pouring water, demonstrating both adaptability and efficiency.

\begin{table}[h!]
\centering
\caption{Pouring performance of different beverage}
\label{diff beverage}
\begin{tabular}{|C{0.82cm} | C{0.82cm} C{0.82cm}| C{0.82cm} C{0.82cm} C{0.82cm} | C{0.82cm}|}
 \hline
 Target & Final & $\sigma$ & P.V. & F.V. & $\sigma$ & Time\\
 {\%} & {\%} & {\%} & {$ml$} & {$ml$} & {$ml$} & {$s$} \\
 \hline
 \multicolumn{7}{|c|}{MTN DEW} \\
 \hline
 40 &  40.6 & 0.22 & 128.7 & 134.9 & 4.03 & 33.79 \\
 60 & 60.8 & 0.22 & 211 & 211.8 & 0.00 & 24.98 \\
 80 & 79.9 & 0.48 & 287 & 285.8 & 1.41 & 22.14 \\
 \hline
 \multicolumn{7}{|c|}{Sprite} \\
 \hline
 40 & 40.4 & 0.24 & 135 & 134.6 & 0.82 & 37.5  \\
 60 & 60.7 & 0.40 & 213.7 & 211.9 & 0.47 & 32.7  \\
 80 & 80.3 & 0.17 & 294.7 & 286.4 & 3.77 & 41.0  \\
 \hline
 \multicolumn{7}{|c|}{Water} \\
 \hline
 40 & 39.9 & 0.14 & 131 & 131.1 & 4.97 & 31.23 \\
 60 & 60.3 & 0.16 & 217 & 214.1 & 3.74 & 23.27 \\
 80 & 80.2 & 0.17 & 295.3 & 292.4 & 0.47 & 17.46 \\
 \hline
\end{tabular}
\end{table}

\subsubsection{Different Target Containers}
Table \ref{diff target container} presents the pouring performance for two different target containers: a measuring cup and a transparent bowl. Due to the large diameter of the transparent bowl, it is not feasible to achieve fill levels of 80\% or 60\% using a single Coke can. As a result, we tested the bowl with target fill levels of 30\%, 35\%, and 40\%. Our pipeline exhibited exceptional performance with the measuring cup, achieving precise fluid levels that closely matched the target percentages. In contrast, the transparent bowl, owing to its larger diameter, presented a unique challenge. As shown in the table, achieving precise fill levels required slightly more time due to the PID controller's slower angle adjustments.

\begin{table}[h!]
\centering
\caption{Pouring performance of different target container}
\label{diff target container}
\begin{tabular}{|C{0.82cm} | C{0.82cm} C{0.82cm}| C{0.82cm} C{0.82cm} C{0.82cm} | C{0.82cm}|}
 \hline
 Target & Final & $\sigma$ & P.V. & F.V. & $\sigma$ & Time\\
 {\%} & {\%} & {\%} & {$ml$} & {$ml$} & {$ml$} & {$s$} \\
 \hline
 \multicolumn{7}{|c|}{Measuring Cup} \\
 \hline
 40 &  41.5  & 0.61 & 129 & 128.7 & 2.16 & 38.67  \\
 60 & 60.2 & 0.37 & 193.7 & 193.1 & 4.71 & 33.05   \\
 80 & 80.7 & 0.22 & 273.7 & 260.7 & 1.89 & 32.17   \\
 \hline
 \multicolumn{7}{|c|}{Transparent Bowl} \\
 \hline
 30 & 29.7 & 0.29 & 134.3 & 137.9 & 8.18 & 52.3  \\
 35 & 35.2 & 0.29 & 210.7 & 210.4 & 13.47 & 76.2  \\
 40 & 39.9 & 0.047 & 274.3 & 267.8 & 2.05 & 99.1  \\
 \hline
\end{tabular}
\end{table}
\subsubsection{Different Camera Locations}
Table \ref{diff location} showcases the pouring performance with different camera locations: ``location 2'' and ``location 3". Notably, at ``location 2", our system consistently achieved accurate pouring results with minimal deviations. The final fluid levels closely matched the target percentages, and our volume estimation method proved highly effective. For instance, at an 80\% target, the predicted volume was 285.7 ml, and the final volume achieved was 278.1 ml. Additionally, pouring times were efficient, ranging from 34.6 seconds to 41.0 seconds for an 80\% target. In contrast, at ``location 3", the system faced slightly greater challenges, as indicated in the table. While it still achieved accurate pouring, the deviations in final fluid levels and pouring times were slightly higher compared to ``location 2".

\begin{table}[h!]
\centering
\caption{Pouring performance of different camera location}
\label{diff location}
\begin{tabular}{|C{0.82cm} | C{0.82cm} C{0.82cm}| C{0.82cm} C{0.82cm} C{0.82cm} | C{0.82cm}|}
 \hline
 Target & Final & $\sigma$ & P.V. & F.V. & $\sigma$ & Time\\
 {\%} & {\%} & {\%} & {$ml$} & {$ml$} & {$ml$} & {$s$} \\
 \hline
 \multicolumn{7}{|c|}{location 2} \\
 \hline
 40 &  40.5 & 0.50 & 129.3 & 126.9 & 3.09 & 34.6\\
 60 & 60.6 & 0.59 & 212.3 & 207.5 & 2.49 & 35.2 \\
 80 & 80.9 & 0.29 & 285.7 & 278.1 & 3.40 & 34.6  \\
 \hline
 \multicolumn{7}{|c|}{location 3} \\
 \hline
 40 & 40.4 & 0.66 & 127.3 & 128.1 & 1.25 & 50.4 \\
 60 & 60.2 & 0.21 & 210.3 & 206.0 & 2.49 & 30.3 \\
 80 & 80.5 & 0.59 & 295.7 & 278.3 & 8.06 & 30.6 \\
 \hline
\end{tabular}
\end{table}

\section{CONCLUSION}

In this paper, we introduce a novel autonomous robotic pipeline designed for real-time pouring using a single RGB input. Our pipeline demonstrates zero-shot robotic pouring capabilities with pre-trained vision models, achieving exceptional accuracy in both pouring control and volume estimation. We validate its performance across various scenarios, pouring liquids with different carbonation and transparency, into diverse transparent containers from different camera positions. Moreover, by integrating ChatGPT, our approach becomes user-friendly for individuals with varying expertise. In this work, we focus only on a single pour into a target container with a fixed location, in future, we aim to expand our pipeline to conduct multi-pour tasks involving multiple target containers in random locations. This extension will challenge our system to handle complex scenarios and interactions, paving the way for more versatile applications. Furthermore, the integration of ChatGPT could be deepened within the pouring pipeline, enabling it to assist in creating task plans for multi-pour scenarios \cite{vemprala_chatgpt_2023}. We believe that our approach represents an exciting direction for the future of robotic applications, especially in the development of household robots.












\printbibliography

@inproceedings{schenck_visual_2017,
	location = {Singapore, Singapore},
	title = {Visual closed-loop control for pouring liquids},
	isbn = {978-1-5090-4633-1},
	doi = {10.1109/ICRA.2017.7989307},
	eventtitle = {2017 {IEEE} International Conference on Robotics and Automation ({ICRA})},
	pages = {2629--2636},
	booktitle = {2017 {IEEE} International Conference on Robotics and Automation ({ICRA})},
	publisher = {{IEEE}},
	author = {Schenck, Connor and Fox, Dieter},
	date = {2017-05},
	langid = {english},
}

@misc{narasimhan_self-supervised_2022,
	title = {Self-supervised Transparent Liquid Segmentation for Robotic Pouring},
	number = {{arXiv}:2203.01538},
	publisher = {{arXiv}},
	author = {Narasimhan, Gautham Narayan and Zhang, Kai and Eisner, Ben and Lin, Xingyu and Held, David},
	date = {2022-03-03},
	langid = {english},
	eprinttype = {arxiv},
	eprint = {2203.01538 [cs]},
}

@inproceedings{liang_robust_2020,
	location = {Las Vegas, {NV}, {USA}},
	title = {Robust Robotic Pouring using Audition and Haptics},
	isbn = {978-1-72816-212-6},
	doi = {10.1109/IROS45743.2020.9340859},
	eventtitle = {2020 {IEEE}/{RSJ} International Conference on Intelligent Robots and Systems ({IROS})},
	pages = {10880--10887},
	booktitle = {2020 {IEEE}/{RSJ} International Conference on Intelligent Robots and Systems ({IROS})},
	publisher = {{IEEE}},
	author = {Liang, Hongzhuo and Zhou, Chuangchuang and Li, Shuang and Ma, Xiaojian and Hendrich, Norman and Gerkmann, Timo and Sun, Fuchun and Stoffel, Marcus and Zhang, Jianwei},
	date = {2020-10-24},
	langid = {english},
}

@article{kennedy_autonomous_2019,
	title = {Autonomous Precision Pouring From Unknown Containers},
	volume = {4},
	issn = {2377-3766, 2377-3774},
	doi = {10.1109/LRA.2019.2902075},
	pages = {2317--2324},
	number = {3},
	journaltitle = {{IEEE} Robotics and Automation Letters},
	shortjournal = {{IEEE} Robot. Autom. Lett.},
	author = {Kennedy, Monroe and Schmeckpeper, Karl and Thakur, Dinesh and Jiang, Chenfanfu and Kumar, Vijay and Daniilidis, Kostas},
	date = {2019-07},
}

@inproceedings{dong_precision_2019,
	location = {Macau, China},
	title = {Precision Pouring into Unknown Containers by Service Robots},
	isbn = {978-1-72814-004-9},
	doi = {10.1109/IROS40897.2019.8967911},
	eventtitle = {2019 {IEEE}/{RSJ} International Conference on Intelligent Robots and Systems ({IROS})},
	pages = {5875--5882},
	booktitle = {2019 {IEEE}/{RSJ} International Conference on Intelligent Robots and Systems ({IROS})},
	publisher = {{IEEE}},
	author = {Dong, Chenyu and Takizawa, Masaru and Kudoh, Shunsuke and Suehiro, Takashi},
	date = {2019-11},
}

@inproceedings{babaians_pournet_2022,
	location = {Kyoto, Japan},
	title = {{PourNet}: Robust Robotic Pouring Through Curriculum and Curiosity-based Reinforcement Learning},
	isbn = {978-1-66547-927-1},
	doi = {10.1109/IROS47612.2022.9981195},
	shorttitle = {{PourNet}},
	eventtitle = {2022 {IEEE}/{RSJ} International Conference on Intelligent Robots and Systems ({IROS})},
	pages = {9332--9339},
	booktitle = {2022 {IEEE}/{RSJ} International Conference on Intelligent Robots and Systems ({IROS})},
	publisher = {{IEEE}},
	author = {Babaians, Edwin and Sharma, Tapan and Karimi, Mojtaba and Sharifzadeh, Sahand and Steinbach, Eckehard},
	date = {2022-10-23},
}

@inproceedings{matl_haptic_2019,
	location = {Macau, China},
	title = {Haptic Perception of Liquids Enclosed in Containers},
	isbn = {978-1-72814-004-9},
	doi = {10.1109/IROS40897.2019.8968528},
	eventtitle = {2019 {IEEE}/{RSJ} International Conference on Intelligent Robots and Systems ({IROS})},
	pages = {7142--7149},
	booktitle = {2019 {IEEE}/{RSJ} International Conference on Intelligent Robots and Systems ({IROS})},
	publisher = {{IEEE}},
	author = {Matl, Carolyn and Matthew, Robert and Bajcsy, Ruzena},
	date = {2019-11},
}

@inproceedings{burns_look_2022,
	location = {Philadelphia, {PA}, {USA}},
	title = {Look and Listen: A Multi-Sensory Pouring Network and Dataset for Granular Media from Human Demonstrations},
	isbn = {978-1-72819-681-7},
	doi = {10.1109/ICRA46639.2022.9812125},
	shorttitle = {Look and Listen},
	eventtitle = {2022 {IEEE} International Conference on Robotics and Automation ({ICRA})},
	pages = {2519--2524},
	booktitle = {2022 International Conference on Robotics and Automation ({ICRA})},
	publisher = {{IEEE}},
	author = {Burns, Alexis and Xiang, Siyuan and Lee, Daewon and Jackel, Larry and Song, Shuran and Isler, Volkan},
	date = {2022-05-23},
}

@inproceedings{piacenza_pouring_2022,
	location = {Philadelphia, {PA}, {USA}},
	title = {Pouring by Feel: An Analysis of Tactile and Proprioceptive Sensing for Accurate Pouring},
	isbn = {978-1-72819-681-7},
	doi = {10.1109/ICRA46639.2022.9811898},
	shorttitle = {Pouring by Feel},
	eventtitle = {2022 {IEEE} International Conference on Robotics and Automation ({ICRA})},
	pages = {10248--10254},
	booktitle = {2022 International Conference on Robotics and Automation ({ICRA})},
	publisher = {{IEEE}},
	author = {Piacenza, Pedro and Lee, Daewon and Isler, Volkan},
	date = {2022-05-23},
}

@inproceedings{reyes-montiel_geometric_2022,
	location = {Lisbon, Portugal},
	title = {A Geometric Approach for Partial Liquids’ Pouring from a Regular Container by a Robotic Manipulator:},
	isbn = {978-989-758-585-2},
	doi = {10.5220/0011321600003271},
	shorttitle = {A Geometric Approach for Partial Liquids’ Pouring from a Regular Container by a Robotic Manipulator},
	eventtitle = {19th International Conference on Informatics in Control, Automation and Robotics},
	pages = {688--694},
	booktitle = {Proceedings of the 19th International Conference on Informatics in Control, Automation and Robotics},
	publisher = {{SCITEPRESS} - Science and Technology Publications},
	author = {Reyes-Montiel, Jeeangh and Marin-Hernandez, Antonio and Hernandez-Mendez, Sergio},
	date = {2022},
}

@misc{zhang_modular_2020,
	title = {A Modular Robotic Arm Control Stack for Research: Franka-Interface and {FrankaPy}},
	shorttitle = {A Modular Robotic Arm Control Stack for Research},
	number = {{arXiv}:2011.02398},
	publisher = {{arXiv}},
	author = {Zhang, Kevin and Sharma, Mohit and Liang, Jacky and Kroemer, Oliver},
	date = {2020-11-04},
	eprinttype = {arxiv},
	eprint = {2011.02398 [cs]},
}

@misc{vaswani_attention_2023,
	title = {Attention Is All You Need},
	number = {{arXiv}:1706.03762},
	publisher = {{arXiv}},
	author = {Vaswani, Ashish and Shazeer, Noam and Parmar, Niki and Uszkoreit, Jakob and Jones, Llion and Gomez, Aidan N. and Kaiser, Lukasz and Polosukhin, Illia},
	date = {2023-08-01},
	eprinttype = {arxiv},
	eprint = {1706.03762 [cs]},
}

@misc{lin_microsoft_2015,
	title = {Microsoft {COCO}: Common Objects in Context},
	shorttitle = {Microsoft {COCO}},
	number = {{arXiv}:1405.0312},
	publisher = {{arXiv}},
	author = {Lin, Tsung-Yi and Maire, Michael and Belongie, Serge and Bourdev, Lubomir and Girshick, Ross and Hays, James and Perona, Pietro and Ramanan, Deva and Zitnick, C. Lawrence and Dollár, Piotr},
	date = {2015-02-20},
	eprinttype = {arxiv},
	eprint = {1405.0312 [cs]},
}

@misc{redmon_you_2016,
	title = {You Only Look Once: Unified, Real-Time Object Detection},
	shorttitle = {You Only Look Once},
	number = {{arXiv}:1506.02640},
	publisher = {{arXiv}},
	author = {Redmon, Joseph and Divvala, Santosh and Girshick, Ross and Farhadi, Ali},
	date = {2016-05-09},
	eprinttype = {arxiv},
	eprint = {1506.02640 [cs]},
}

@misc{vemprala_chatgpt_2023,
	title = {{ChatGPT} for Robotics: Design Principles and Model Abilities},
	shorttitle = {{ChatGPT} for Robotics},
	number = {{arXiv}:2306.17582},
	publisher = {{arXiv}},
	author = {Vemprala, Sai and Bonatti, Rogerio and Bucker, Arthur and Kapoor, Ashish},
	date = {2023-07-19},
	eprinttype = {arxiv},
	eprint = {2306.17582 [cs]},
}

@misc{liu_keypose_2020,
	title = {{KeyPose}: Multi-View 3D Labeling and Keypoint Estimation for Transparent Objects},
	shorttitle = {{KeyPose}},
	number = {{arXiv}:1912.02805},
	publisher = {{arXiv}},
	author = {Liu, Xingyu and Jonschkowski, Rico and Angelova, Anelia and Konolige, Kurt},
	date = {2020-05-18},
	eprinttype = {arxiv},
	eprint = {1912.02805 [cs]},
}

@misc{sajjan_cleargrasp_2019,
	title = {{ClearGrasp}: 3D Shape Estimation of Transparent Objects for Manipulation},
	url = {http://arxiv.org/abs/1910.02550},
	shorttitle = {{ClearGrasp}},
	number = {{arXiv}:1910.02550},
	publisher = {{arXiv}},
	author = {Sajjan, Shreeyak S. and Moore, Matthew and Pan, Mike and Nagaraja, Ganesh and Lee, Johnny and Zeng, Andy and Song, Shuran},
	date = {2019-10-14},
	eprinttype = {arxiv},
	eprint = {1910.02550 [cs, eess]},
}

@misc{zhu_rgb-d_2021,
	title = {{RGB}-D Local Implicit Function for Depth Completion of Transparent Objects},
	number = {{arXiv}:2104.00622},
	publisher = {{arXiv}},
	author = {Zhu, Luyang and Mousavian, Arsalan and Xiang, Yu and Mazhar, Hammad and van Eenbergen, Jozef and Debnath, Shoubhik and Fox, Dieter},
	date = {2021-04-01},
	eprinttype = {arxiv},
	eprint = {2104.00622 [cs]},
}

@misc{schenck_towards_2016,
	title = {Towards Learning to Perceive and Reason About Liquids},
	number = {{arXiv}:1608.00887},
	publisher = {{arXiv}},
	author = {Schenck, Connor and Fox, Dieter},
	date = {2016-08-02},
	eprinttype = {arxiv},
	eprint = {1608.00887 [cs]},
}

@misc{chen_rethinking_2017,
	title = {Rethinking Atrous Convolution for Semantic Image Segmentation},
	number = {{arXiv}:1706.05587},
	publisher = {{arXiv}},
	author = {Chen, Liang-Chieh and Papandreou, George and Schroff, Florian and Adam, Hartwig},
	date = {2017-12-05},
	eprinttype = {arxiv},
	eprint = {1706.05587 [cs]},
}

@misc{ronneberger_u-net_2015,
	title = {U-Net: Convolutional Networks for Biomedical Image Segmentation},
	shorttitle = {U-Net},
	number = {{arXiv}:1505.04597},
	publisher = {{arXiv}},
	author = {Ronneberger, Olaf and Fischer, Philipp and Brox, Thomas},
	date = {2015-05-18},
	eprinttype = {arxiv},
	eprint = {1505.04597 [cs]},
}

@misc{long_fully_2015,
	title = {Fully Convolutional Networks for Semantic Segmentation},
	number = {{arXiv}:1411.4038},
	publisher = {{arXiv}},
	author = {Long, Jonathan and Shelhamer, Evan and Darrell, Trevor},
	date = {2015-03-08},
	eprinttype = {arxiv},
	eprint = {1411.4038 [cs]},
}

@article{eppel_computer_2020,
	title = {Computer Vision for Recognition of Materials and Vessels in Chemistry Lab Settings and the Vector-{LabPics} Data Set},
	volume = {6},
	issn = {2374-7943, 2374-7951},
	doi = {10.1021/acscentsci.0c00460},
	pages = {1743--1752},
	number = {10},
	journaltitle = {{ACS} Central Science},
	shortjournal = {{ACS} Cent. Sci.},
	author = {Eppel, Sagi and Xu, Haoping and Bismuth, Mor and Aspuru-Guzik, Alan},
	date = {2020-10-28},
	langid = {english},
}

@misc{eppel_computer_2021,
	title = {Computer vision for liquid samples in hospitals and medical labs using hierarchical image segmentation and relations prediction},
	number = {{arXiv}:2105.01456},
	publisher = {{arXiv}},
	author = {Eppel, Sagi and Xu, Haoping and Aspuru-Guzik, Alan},
	date = {2021-05-04},
	eprinttype = {arxiv},
	eprint = {2105.01456 [cs]},
}

@software{jocher_yolo_2023,
	title = {{YOLO} by Ultralytics},
	version = {8.0.0},
	author = {Jocher, Glenn and Chaurasia, Ayush and Qiu, Jing},
	date = {2023-01},
}

@misc{xian_fluidlab_2023,
	title = {{FluidLab}: A Differentiable Environment for Benchmarking Complex Fluid Manipulation},
	shorttitle = {{FluidLab}},
	number = {{arXiv}:2303.02346},
	publisher = {{arXiv}},
	author = {Xian, Zhou and Zhu, Bo and Xu, Zhenjia and Tung, Hsiao-Yu and Torralba, Antonio and Fragkiadaki, Katerina and Gan, Chuang},
	date = {2023-03-04},
	eprinttype = {arxiv},
	eprint = {2303.02346 [cs]},
}

@misc{shah_airsim_2017,
	title = {{AirSim}: High-Fidelity Visual and Physical Simulation for Autonomous Vehicles},
	shorttitle = {{AirSim}},
	number = {{arXiv}:1705.05065},
	publisher = {{arXiv}},
	author = {Shah, Shital and Dey, Debadeepta and Lovett, Chris and Kapoor, Ashish},
	date = {2017-07-18},
	eprinttype = {arxiv},
	eprint = {1705.05065 [cs]},
}

@misc{xie_segmenting_2020,
	title = {Segmenting Transparent Objects in the Wild},
	number = {{arXiv}:2003.13948},
	publisher = {{arXiv}},
	author = {Xie, Enze and Wang, Wenjia and Wang, Wenhai and Ding, Mingyu and Shen, Chunhua and Luo, Ping},
	date = {2020-08-01},
	eprinttype = {arxiv},
	eprint = {2003.13948 [cs]},
}

@misc{xie_segmenting_2021,
	title = {Segmenting Transparent Object in the Wild with Transformer},
	number = {{arXiv}:2101.08461},
	publisher = {{arXiv}},
	author = {Xie, Enze and Wang, Wenjia and Wang, Wenhai and Sun, Peize and Xu, Hang and Liang, Ding and Luo, Ping},
	date = {2021-02-23},
	eprinttype = {arxiv},
	eprint = {2101.08461 [cs]},
}

@article{li_graph_2022,
	title = {Graph neural network-accelerated Lagrangian fluid simulation},
	volume = {103},
	issn = {00978493},
	doi = {10.1016/j.cag.2022.02.004},
	pages = {201--211},
	journaltitle = {Computers \& Graphics},
	shortjournal = {Computers \& Graphics},
	author = {Li, Zijie and Farimani, Amir Barati},
	date = {2022-04},
	langid = {english},
}

@misc{dikshit_robochop_2023,
	title = {{RoboChop}: Autonomous Framework for Fruit and Vegetable Chopping Leveraging Foundational Models},
	shorttitle = {{RoboChop}},
	number = {{arXiv}:2307.13159},
	publisher = {{arXiv}},
	author = {Dikshit, Atharva and Bartsch, Alison and George, Abraham and Farimani, Amir Barati},
	date = {2023-07-24},
	eprinttype = {arxiv},
	eprint = {2307.13159 [cs]},
}

@misc{do_accurate_2018,
	title = {Accurate Pouring with an Autonomous Robot Using an {RGB}-D Camera},
	number = {{arXiv}:1810.03303},
	publisher = {{arXiv}},
	author = {Do, Chau and Burgard, Wolfram},
	date = {2018-10-08},
	eprinttype = {arxiv},
	eprint = {1810.03303 [cs]},
}


\end{document}